\documentclass[letterpaper, 10 pt, journal, twoside]{ieeetran} 
\usepackage[utf8]{inputenc}
\usepackage[T1]{fontenc}
\usepackage{amsmath,amsfonts}
\usepackage{amssymb}
\usepackage[normalem]{ulem}
\usepackage{textcomp}
\usepackage{url}
\usepackage{verbatim}
\usepackage{graphicx}
\usepackage{graphics} 
\usepackage{mathtools}
\usepackage{booktabs}
\usepackage{adjustbox}
\usepackage{stfloats}
\usepackage[breaklinks=true,colorlinks,citecolor=blue]{hyperref}
\usepackage{makecell}
\usepackage{multirow}
\usepackage{balance}
\usepackage{rotating}
\usepackage[linesnumbered,ruled,vlined,commentsnumbered]{algorithm2e}
\usepackage{footnote}
\usepackage[dvipsnames]{xcolor}
\usepackage{tabularray}
\usepackage{pifont}

\newcommand{\red}[1]{{\color{Red}{#1}}}

\newcommand{\blue}[1]{{\color{MidnightBlue}{#1}}}

\newcommand{\green}[1]{{\color{Green}{#1}}}

\newcommand{\orange}[1]{{\color{Orange}{#1}}}

\begin{document}

\title{\texttt{\textbf{\green{InLiER:}}} Learning-Free Heterogeneous LiDAR Place Recognition via Intermediate Mixed-Radix Structural Keypoint Tokenization}

\makeatletter
\newcommand*\titleheader[1]{\gdef\@titleheader{#1}}
\AtBeginDocument{%
  \let\st@red@title\@title
  \def\@title{%
    \vskip-2.0em
    \bgroup\normalfont\small\centering\@titleheader\par\egroup
    \vskip0.5em\st@red@title}
}
\makeatother

\titleheader{Please cite as: N. Stathoulopoulos and G. Nikolakopoulos, "InLiER: Learning-Free Heterogeneous LiDAR Place Recognition via Intermediate Mixed-Radix Structural Keypoint Tokenization," in IEEE Robotics and Automation Letters, vol. 11, no. 10, pp. 11275-11282, Oct. 2026, doi: 10.1109/LRA.2026.3723737.}
%


\author{Nikolaos Stathoulopoulos and George Nikolakopoulos%

\thanks{This paper was supported in part and received funding from the European Union’s Horizon Europe Research and Innovation Programme under the Grant Agreement No. 101138451.}%
\thanks{The authors are with the Robotics and AI Group, Department of Computer, Electrical and Space Engineering, Lule\r{a} University of Technology, 971 87 Lule\r{a}, Sweden. \textit{Corresponding Author's e-mail:} \tt\footnotesize niksta@ltu.se}%
\thanks{Additional material and the source code of this letter are available at our github repo: \textbf{\texttt{\href{https://github.com/LTU-RAI/InLiER.git}{https://github.com/LTU-RAI/InLiER.git}}}.}
}



\maketitle
\begin{abstract}
LiDAR place recognition supports loop closure, relocalization, and multi-agent map management.
As robotic platforms increasingly combine LiDARs with different fields of view, resolutions, and scanning patterns, existing descriptors degrade because they are tightly coupled to sensor-specific characteristics.
We present InLiER, a learning-free pipeline based on an intermediate tokenization step. Height-sliced keypoints from structural elements receive mixed-radix token IDs encoding height, radial distance, local shape, and azimuth from local 3D geometry, in a compact sub-2KB representation. The same vocabulary is reorganized across three retrieval stages: height-ceiling histogram intersection for fast rotation-invariant shortlisting, binary bitmask alignment for yaw estimation and reranking, and token-guided geometric verification for 6-DoF pose estimation.
InLiER achieves state-of-the-art performance on the HeLiPR dataset and in real-world field experiments, among modern handcrafted methods and outperforms the learning-based baseline on most cross-sensor configurations.
\end{abstract}

\vspace{-10pt}

\section{Introduction} \label{sec:introduction}

\IEEEPARstart{L}{iDAR}-based place recognition~(LiPR) is essential for loop closure in SLAM~\cite{loopclosures}, relocalization~\cite{reloc}, and multi-agent map management~\cite{stathoulopoulos_frame_2024}, and offers geometric consistency across illumination and weather where vision-based alternatives degrade~\cite{yuan2024btc}.
Modern platforms increasingly combine heterogeneous LiDARs: spinning sensors with $360^\circ$ coverage (e.g.\ Ouster), solid-state units with a narrow non-repetitive field-of-view (FoV) (e.g.\ Livox Avia, $70^\circ\!\times\!77^\circ$), and FMCW sensors (e.g.\ Aeva Aeries\,II, $120^\circ\!\times\!20^\circ$). 
These sensors differ in FoV, resolution (6--128 channels), and scanning pattern, so two scans of the same place may share only partial spatial overlap~\cite{minwoo2024helipr}.
This heterogeneity exposes two orthogonal failure modes.
\emph{(a) Density variation:} the same surface sampled by a 16-channel sensor yields only a handful of points where a 128-channel device produces dozens, and density-dependent descriptors conflate this with true geometric differences.
\emph{(b) FoV asymmetry:} a $70^\circ$ solid-state and a $360^\circ$ spinning sensor observe fundamentally different scene subsets from the same pose. 
Descriptors that fill a fixed polar grid then read unobserved directions as geometric emptiness, injecting a structured false signal into retrieval.
Under a $70^\circ$ FoV, over $80\%$ of such a descriptor is vacant~\cite{kim2024solid}.
Existing methods address only subsets of these requirements.
BEV descriptors like Scan Context++~\cite{kim2022scpp} and RING++~\cite{Xu2023ringpp} assume $360^\circ$ coverage and collapse under narrow FoV.
SOLiD~\cite{kim2024solid} handles narrow FoV via spatial organization but has not been validated for cross-sensor matching and degrades under density variation.
Learning-based HeLiOS~\cite{jung2025helios} achieves the best heterogeneous retrieval to date. 
However, its overlap-based supervision optimizes for 3D point cloud overlap rather than metric pose proximity, so top retrievals can be substantially displaced from the query. 
This undermines the role of place recognition as spatial initialization for ICP-based metric refinement, which fails under large translational offsets~\cite{stathoulopoulos_frame_2024, yuan2024btc}.
Moreover, HeLiOS yields only a retrieval descriptor with no intermediate structure to repurpose (e.g. for geometric verification, pose estimation etc.)
\begin{figure}[!t]
    \centering
    \includegraphics[width=1.0\linewidth]{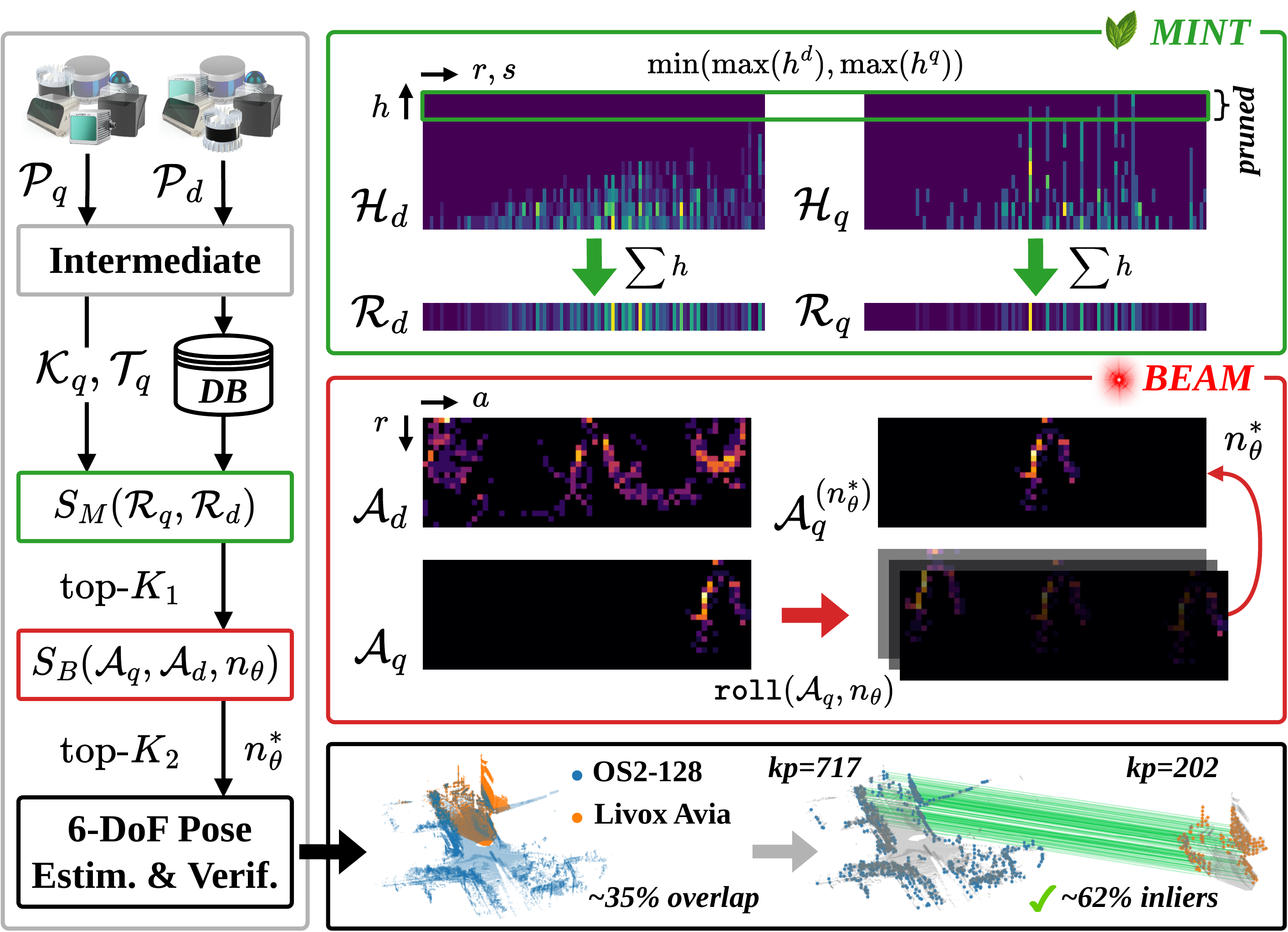}
    \setlength{\abovecaptionskip}{-14pt}
    \caption{Overview of \green{\textbf{\texttt{InLiER}}}. Heterogeneous input scans are mapped to an intermediate representation of keypoints and token IDs. The same token set is progressively reorganized across three retrieval stages: MINT for rotation-invariant shortlisting, BEAM for yaw estimation via binary bitmask alignment, and a final token-guided verification with 6-DoF pose estimation. Despite different scanning patterns and FoVs, the keypoints between the Ouster OS2-128 and Livox Avia yield $\sim\!62\%$ inliers, showcasing cross-sensor robustness.}
    \label{fig:overview}
    \vspace{-18pt}
\end{figure}

Both failure modes share a common cause: descriptors built directly on raw points conflate scene content with sensor sampling.
Within this letter, we introduce InLiER (\textbf{In}termediate \textbf{Li}DAR \textbf{E}ncoding for \textbf{R}etrieval), a learning-free global localization pipeline that breaks this coupling by inserting an intermediate representation between the scan and the descriptor.
This representation consists of height-sliced keypoints extracted from structural elements, each mapped to a compact mixed-radix token.
Since tokens depend on what the sensor sees rather than how it sees it, a narrow-FoV or sparse LiDAR produces fewer tokens instead of a corrupted signal. Thus, descriptor quality degrades gracefully under FoV and resolution disparities.
The same token set is then progressively reorganized across three retrieval stages: rotation-invariant shortlisting, yaw-aware reranking, and geometric verification, yielding a full 6-DoF (degrees-of-freedom) pose estimate from a single intermediate representation, as seen in Fig.~\ref{fig:overview}.

Our contributions are:
(a) A learning-free intermediate LiDAR representation that maps heterogeneous scans into a shared vocabulary of height-sliced structural keypoints and mixed-radix tokens, reducing dependence on sensor FoV, resolution, and scanning pattern while remaining compact for storage and transmission;
(b) A unified multi-stage localization pipeline 
that reorganizes this representation on the fly for each query-database pair. It includes height-ceiling pruning for mutually supported vertical FoV matching. Retrieval, alignment, and pose estimation are all derived from a single representation;
(c) InLiER achieves state-of-the-art performance among handcrafted methods and outperforms the learning-based baseline on most cross-sensor configurations both in HeLiPR and in a real-world field experiment.
It also retrieves spatially close candidates rather than merely high-overlap matches, supporting reliable downstream metric refinement.
\section{Related Work} \label{sec:related-work}

We focus on LiPR methods that address heterogeneity in point density or FoV; for a broader taxonomy, we refer the reader to~\cite{Luo2024pcpcsurvey}. 
Methods assuming homogeneous~$360^\circ$ sensors are discussed only insofar as they motivate InLiER's design.

The dominant family of handcrafted descriptors encodes scans as bird's-eye-view (BEV) projections.
Scan Context++~\cite{kim2022scpp} uses complementary polar and Cartesian sub-descriptors for pseudo-invariance to yaw rotation and lateral translation, with efficient \emph{k}-D tree retrieval.
RING++~\cite{Xu2023ringpp} extends this paradigm analytically: successive Radon and Fourier transforms over a multichannel BEV yield a representation with provable roto-translation invariance and globally convergent pose solvers. 
It is the first learning-free framework to jointly address retrieval and 3-DoF localization on a sparse scan.
Both share a structural limitation in heterogeneous settings: a fixed polar grid treats unobserved directions as geometric emptiness, injecting a structured false signal rather than gracefully reducing information content.
This failure has motivated more selective encoding strategies. 
BTC~\cite{yuan2024btc} abandons the global grid in favor of local geometric primitives. Keypoints from detected plane boundaries form triangles whose sorted side lengths are rigidly invariant, enabling $\mathcal{O}(1)$ hash-table retrieval and direct 6-DoF pose recovery. Matches are refined by a binary height-encoding string at each vertex.
Operating directly in 3D grants greater sensor adaptability, but heterogeneous FoVs alter observable plane cross-sections and shift keypoint locations, while horizontal~FoV~(HFoV) asymmetry changes which triangles can be formed, ambiguities the descriptor cannot resolve.
SOLiD~\cite{kim2024solid} explicitly targets restricted-FoV scenarios via range-elevation and azimuth-elevation histograms reweighted by vertical point dominance, so that sparse observations reduce descriptor density gracefully rather than catastrophically.
DVMM~\cite{duan_dvmm_2025} pursues a similar dual-view philosophy, combining a Gaussian-weighted azimuthal descriptor with a height-filtered binary BEV cross-section and demonstrating cross-modal retrieval across several LiDAR types.
Both, however, embed implicit shared-FoV assumptions.
SOLiD parameterizes its elevation binning per sensor and was validated only with query and database from the same LiDAR, while DVMM's Gaussian weighting presumes a common equatorial vertical~FoV~(VFoV) region, an assumption that breaks down when sensor orientations or FoV extents differ substantially.
HeLiOS~\cite{jung2025helios} is the only method to date that explicitly addresses all three heterogeneity axes: FoV, resolution, and scanning pattern, within a single framework. 
It learns a shared descriptor using sparse-convolutions, local spherical-window attention, and optimal-transport feature aggregation. It also replaces distance-based sample mining with a 3D overlap criterion, which remains meaningful even when narrow-FoV scans from the same pose do not spatially overlap.
HeLiOS achieves the best heterogeneous retrieval on HeLiPR~\cite{minwoo2024helipr}, but its 3D overlap supervision biases the descriptor toward scan-overlap rather than proximity, so retrievals may be distant from the query and thus poor initializations for downstream pose refinement.
The descriptor is also limited to retrieval, with no yaw or pose estimation stages, leaving any complete pipeline to be assembled from separately designed components.
Beyond descriptor design, two recent pipelines share InLiER's structure of ground alignment, reduced-DoF estimation, and geometric verification. 
MapClosures~\cite{gupta2024efficiently} ground-align accumulated local maps and match image features on density-preserving BEV projections, agnostic to scanning pattern and platform motion.
TreeLoc~\cite{jung_treeloc_2026} aligns tree-stem landmarks via their axes and couples histogram retrieval with triangle matching and two-step verification for 6-DoF localization in forests.

InLiER occupies a complementary position: it is learning-free and addresses both FoV asymmetry and density variation through an explicit intermediate tokenization that decouples place description from sensor sampling. A single compact token vocabulary is reorganized across all stages, with pairwise height-ceiling pruning for explicit VFoV handling, rather than relying on dense BEV imagery~\cite{gupta2024efficiently} or domain-specific landmarks~\cite{jung_treeloc_2026}. From this same representation it produces fast retrieval, verification, and 6-DoF pose estimation.

\section{Methodology} \label{sec:methodology}
Given an input point cloud, InLiER first extracts an intermediate representation consisting of keypoints and their associated token IDs. In contrast to conventional pipelines that encode each scan into a single descriptor through a predetermined spatial organization, InLiER retains this intermediate representation and reorganizes it during querying.
Two complementary mechanisms exploit this flexibility. First, within each stage, the descriptor is conditioned on the query-candidate pair: unsupported height levels are pruned and the overlapping VFoV is matched implicitly, enabling robust comparison between sensors with unknown and different configurations. Second, across stages, the same token set is projected onto progressively richer structures: the early stage produces compact rotation-invariant descriptors for fast shortlisting, the next stage restores directional information for selective scoring such as yaw alignment, and the final stage operates on token correspondences for verification and 6-DoF pose estimation.

\subsection{Preprocessing} \label{subsec:preprocessing}

Following recent works~\cite{yuan2024btc,duan_dvmm_2025}, we adopt a submap as the input unit. Each submap is formed by accumulating $N$ consecutive LiDAR scans to reduce sparsity and improve structural completeness under heterogeneous scanning patterns and resolutions. Let the resulting submap at time $t$ be denoted by $\mathcal{P}_t\!=\!\{\mathbf{p}_i\!\in\!\mathbb{R}^3\}_{i=1}^{M_t}$. We first apply voxel downsampling with voxel size $v_s$ and retain only points within $[R_{\min}, R_{\max}]$, partially normalizing point density and sensing range across sensors.
To handle differences in mounting height and orientation, we then establish a gravity-aligned frame following~\cite{duan_dvmm_2025}. We estimate the dominant ground plane $\boldsymbol{\pi}_g\!=\![\mathbf{n}_g^\top,d_g]^\top$ from $\mathcal{P}_t$ via RANSAC~\cite{RANSAC} and compute a rigid transformation $\mathbf{T}_\text{ground}\!\in\!SE(3)$ that aligns the ground normal with the vertical axis and places the ground plane at $z\!=\!0$, yielding the aligned submap $\bar{\mathcal{P}}_t$, as shown in Fig.~\ref{fig:keypoints}\,(a,\,b$_1$,\,b$_2$). 
%
After alignment, all submaps share a common height baseline, enabling consistent height slices for the subsequent keypoint extraction and providing the basis for handling VFoV mismatch. Finally, the points are cropped to a fixed region of interest, $|x|,|y|\!\leqslant\!L$ and $z\!\in\![z_{\min}, z_{\max}]$, where $L$ is the spatial half-extent.

\vspace{-9pt}
\subsection{Intermediate LiDAR Encoding} \label{subsec:intermediate_encoding}
Given the preprocessed submap $\bar{\mathcal{P}}_t$, InLiER constructs an intermediate representation consisting of \textit{height-sliced keypoints} and their associated \textit{token IDs}. The keypoints form a sparse, height-indexed structural scaffold, and each is assigned a local shape class for additional discriminability. Each keypoint is then mapped one-to-one to a discrete token ID that encodes its height, radial position, shape class, and azimuth as bin indices. The resulting representation is compact and fully decomposable, supporting stage-specific descriptor construction during retrieval without revisiting the raw points.

\subsubsection{Height-Sliced Keypoint Extraction} \label{subsubsec:keypoints}
\begin{figure}[!t]
    \centering
    \includegraphics[width=1.\columnwidth]{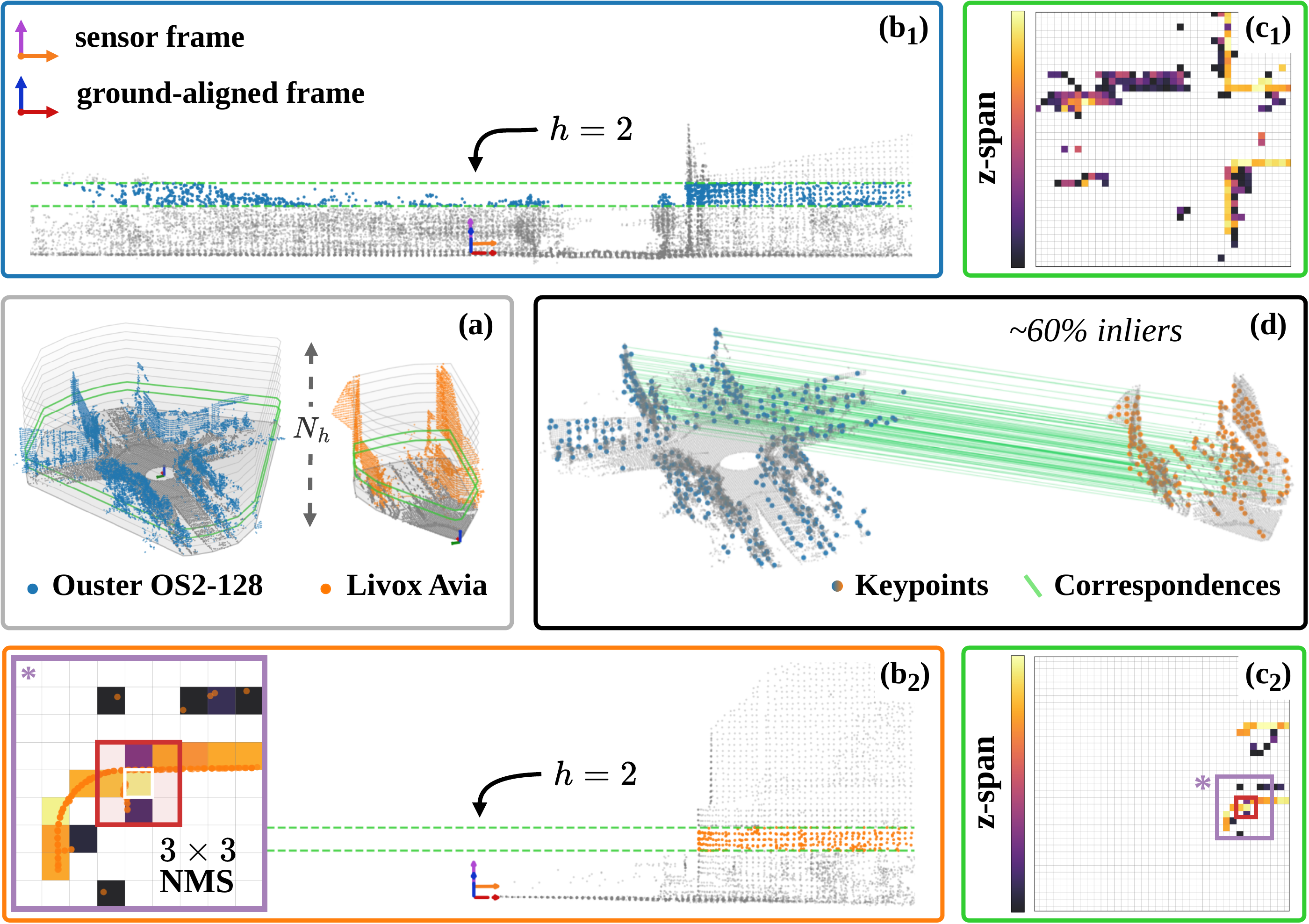}
    \setlength{\abovecaptionskip}{-14pt}
    \caption{Overview of InLiER keypoint extraction
    (a) Both point clouds after RANSAC ground-plane estimation (gray points) and partitioning into $N_h$ height slices. (b$_1$, b$_2$) Side views in the ground-aligned frame with slice $h\!=\!2$ highlighted (blue: Ouster, orange: Livox); the sensor-frame and ground-aligned axes are shown at the origin. (c$_1$, c$_2$) Top-down BEV $z$-span intensity images for the corresponding slice; the inset in (b$_2$, c$_2$) details the $w \times w$ Non-Maximum Suppression (NMS) window. (d) Extracted keypoints from both sensors with correspondences ($\sim\!60\%$~inliers), demonstrating cross-sensor repeatability despite vastly different scanning patterns and FoVs.} \label{fig:keypoints}
    \vspace{-15pt}
\end{figure}

The aligned vertical range $[z_{\min},z_{\max}]$ is partitioned into $N_h$ uniform height slices with edges $z^{(h)}\!=\!z_{\min}\!+\!h\,\Delta z$, where $\Delta z\!=\!(z_{\max}\!-\!z_{\min})/N_h$ and $h\!=\!0,\dots,N_h$, as shown in Fig.~\ref{fig:keypoints}\,(a). For each slice $h$, we collect the subset $\bar{\mathcal{P}}_t^{(h)}\!=\!\{\bar{\mathbf{p}}\!\in\!\bar{\mathcal{P}}_t\!:\!z^{(h)}\!\leqslant\!\bar{p}_z\!<\!z^{(h+1)}\}$ and project it onto a BEV grid of cell size $c_s\!=\!2v_s$ over the fixed region of interest, yielding an image of size $H\!\times\!W$ with $H\!=\!W\!=\!2L/c_s$.
An example is given in Fig.~\ref{fig:keypoints}\,(b$_1$,\,c$_1$,\,b$_2$,\,c$_2$).
For each cell $(x,y)$, we define a \emph{saliency} value from the vertical span ($z$-span) of its contained points as:
\vspace{-2.5pt}
\begin{equation}
    I^{(h)}(x,y) = \max_{\bar{\mathbf{p}}\in\mathcal{C}_{x,y}^{(h)}} \bar{p}_z - \min_{\bar{\mathbf{p}}\in\mathcal{C}_{x,y}^{(h)}} \bar{p}_z,
\end{equation}
where $\mathcal{C}_{x,y}^{(h)} \subset \bar{\mathcal{P}}_t^{(h)}$ is the set of points falling into cell $(x,y)$, and $I^{(h)}(x,y)\!=\!0$ for empty cells. This quantity emphasizes vertically expressive structures while being less sensitive to absolute point count than density-based statistics and more locally descriptive than maximum-height selection.
Keypoints are detected as local maxima of $I^{(h)}$ under Non-Maximum Suppression (NMS) over a $w\times w$ neighborhood, subject to a minimum saliency $\tau_I\!=\!\Delta z/2$, as shown in Fig.~\ref{fig:keypoints}\,(b$_2$,\,c$_2$). Within each slice, the top-$K_h$ cells by saliency are retained; a global cap $K_{\max}$ is then applied by re-ranking the retained cells across all slices by $I^{(h)}(x,y)$. 
For each retained cell, the keypoint position $\bar{\mathbf{p}}^c \!\in\!\mathbb{R}^3$ is defined as the arithmetic mean of the points in $\mathcal{C}^{(h)}_{x,y}$.
The resulting keypoint set $\mathcal{K}_t\!=\!\{\bar{\mathbf{p}}^c_k\}_{k=1}^{K_t}$ is a sparse, height-indexed collection of structurally salient anchors, with an example shown in Fig.~\ref{fig:keypoints}\,(d).

\subsubsection{Local Shape Classification} \label{subsubsec:shape}

To improve discriminability beyond spatial binning alone, each keypoint center $\bar{\mathbf{p}}^c_k$ is assigned a local structural class from PCA~\cite{pca} over a spherical neighborhood $\mathcal{N}_k\!=\!\{\mathbf{p}\!\in\!\mathcal{P}_t\!:\!\|\mathbf{p}-\bar{\mathbf{p}}^c_k\|\!\leqslant\!r_\text{PCA}\}$. If $|\mathcal{N}_k|\!<\!n_{\min}$, the keypoint is directly assigned to the scatter class. Otherwise, we compute the covariance matrix $\boldsymbol{\Sigma}_k$. 
Let $\lambda_1\!\geqslant\!\lambda_2\!\geqslant\!\lambda_3\!\geqslant\!0$ be the eigenvalues of $\boldsymbol{\Sigma}_k$, with corresponding eigenvectors $\mathbf{e}_1,\mathbf{e}_2,\mathbf{e}_3$. 
We then compute the standard linearity, planarity, and scatter scores~\cite{weinmann2015semantic} as: $L_k\!=\!(\lambda_1-\lambda_2)/\lambda_1,\, P_k\!=\!(\lambda_2-\lambda_3)/\lambda_1, \,S_k\!=\!\lambda_3/\lambda_1$, respectively.
Each keypoint receives a single shape label $s_b^{(k)}$. The base type is assigned based on $\arg\max(L_k, P_k, S_k)$. 
Linear and planar types are refined by inclination relative to gravity, measured from the dominant direction $\mathbf{e}_1$ (linear) or surface normal $\mathbf{e}_3$ (planar):$\phi_k\!=\!\arccos\!\left(|\mathbf{e}_z^\top \mathbf{n}_k|\right), \;\phi_k\!\in\![0,\pi/2]$,
where $\mathbf{n}_k$ is this selected axis. Quantizing $\phi_k$ uniformly into $n_\phi$ bins separates, e.g., vertical poles from horizontal edges and ground from walls; scatter is left undivided. Concatenating the base type with its inclination bin gives $s_b^{(k)}\!\in\!\{0,\dots\!,N_s\!-\!1\}$, with $N_s\!=\!2n_\phi + 1$ the total number of shape classes.
\begin{figure}[!t]
    \centering
    \includegraphics[width=1.\columnwidth]{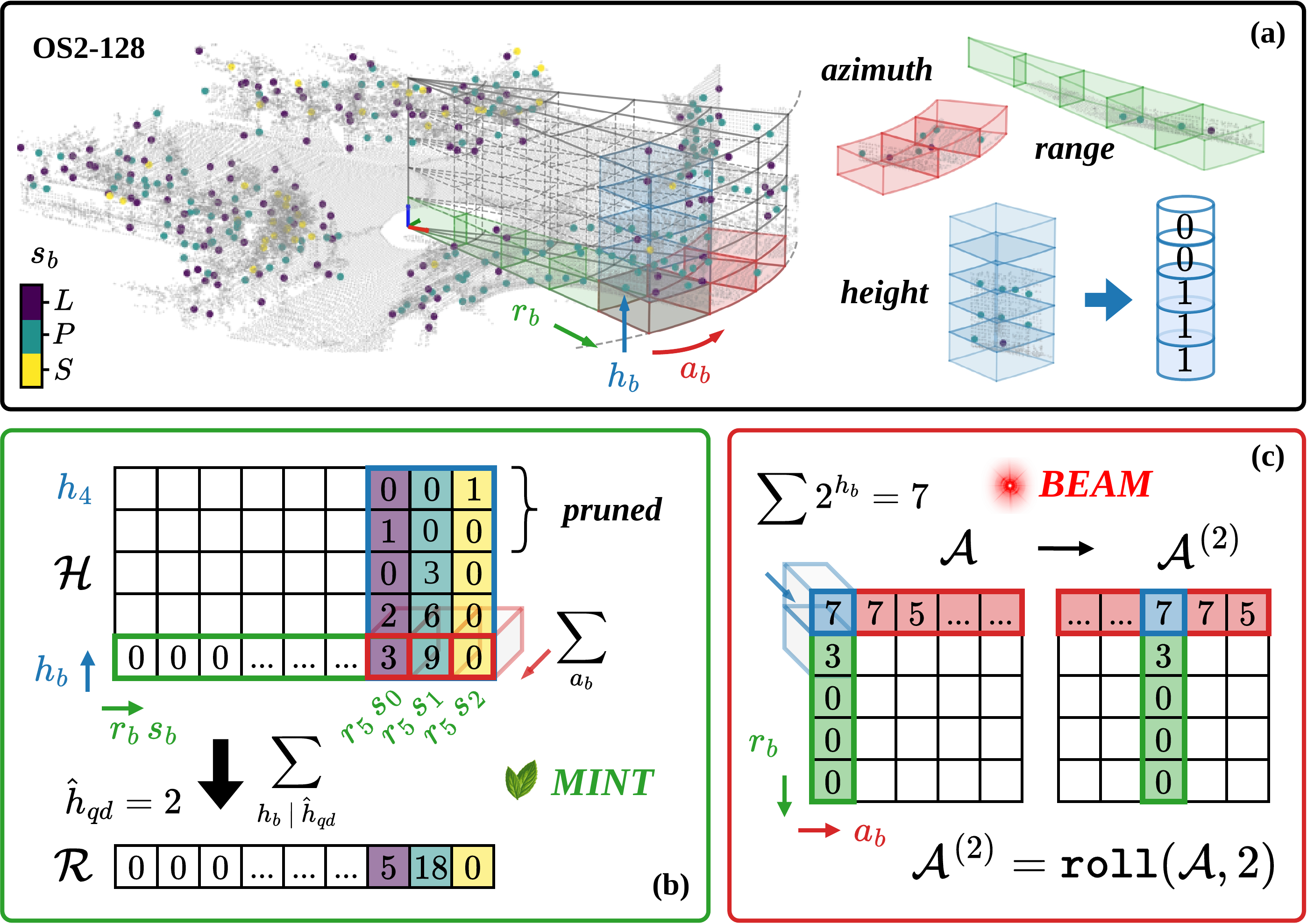}
    \setlength{\abovecaptionskip}{-14pt}
    \caption{Overview of the keypoint-based spatial organization. (a) The input point cloud is discretized into radial, azimuthal, and height bins, with keypoints assigned to shape classes $s_b$. (b) In the Minimum-ceiling Intersection (MINT) retrieval stage, azimuth is marginalized to form $\mathcal{H}$. Height bins above the query-database minimum ceiling $\hat{h}_{qd}$ are pruned, and the remaining height columns are summed to produce the retrieval key $\mathcal{R}$. (c) In the Binary-Elevation Azimuth Matching (BEAM) alignment stage, azimuth is retained to construct the alignment key $\mathcal{A}$. Circular shifts of $\mathcal{A}$ are compared to recover the yaw shift that maximizes the overlap of occupied height bits.}
    \vspace{-15pt}
    \label{fig:key_construction}
\end{figure}

\subsubsection{Token Encoding} \label{sec:tokens}
Each retained keypoint is then mapped $1$--$1$ to a discrete token, yielding a sparse cylindrical-coordinate discretization of the scene as depicted in Fig.~\ref{fig:key_construction}\,(a). Each keypoint is indexed by four bin quantities: a height bin $h_b^{(k)}\!\in\!\{0,\dots\!,\!N_h\!-\!1\}$ given by the slice index; a radial bin $r_b^{(k)}\!\in\!\{0,\dots\!,\!N_r\!-\!1\}$ from uniform quantization of $r_k\!=\!\sqrt{x_k^2+y_k^2}$ over $[0,R_{\max}]$; a shape bin $s_b^{(k)}$ from Sec.~\ref{subsubsec:shape}; and an azimuth bin $a_b^{(k)}\!\in\!\{0,\dots\!,N_a\!-\!1\}$ from uniform quantization of $\theta_k\!=\!\mathrm{atan2}(y_k,x_k)$ over $[0,2\pi)$. 
These indices are packed into a mixed-radix token ID as:
\begin{equation}
    t_{\mathrm{id}}^{(k)} = \big((h_b^{(k)}N_r + r_b^{(k)})N_s + s_b^{(k)}\big)N_a + a_b^{(k)},
\end{equation}
The packing is bijective: any token ID decomposes back into its constituent $(h_b, r_b, s_b, a_b)$ bins, so the token set can be reorganized during retrieval to form different descriptors and matching spaces, e.g. suppressing azimuth for rotation-invariant retrieval or pruning unsupported height bins for VFoV-aware matching.
We denote the resulting token set as $\mathcal{T}_t\!=\!\{t_{\mathrm{id}}^{(k)}\}_{k=1}^{K_t}$. Thus, $\texttt{\textbf{InLiER}}_{\,t}\!=\!(\mathcal{K}_t,\mathcal{T}_t)$ defines the proposed intermediate representation. For retrieval, at most $K_{\max}$ tokens are stored or transmitted, yielding a compact representation of size $K_{\max}\!\times\!1$. If geometric verification or pose estimation is required, the keypoints are retained along their tokens as $(x,y,z,t_{\mathrm{id}})$, yielding a $K_{\max}\!\times\!4$ representation.

\vspace{-8pt}
\subsection{Multi-Stage Place Recognition} \label{subsec:matching}

Given 
$\texttt{\textbf{InLiER}}_{\,t}$, 
LiPR 
proceeds in three progressive stages, each reorganizing the same token set to address a distinct source of heterogeneity. 
First, MINT~(Sec.\,\ref{subsubsec:mint}) performs efficient database-wide shortlisting via a rotation-invariant, height-ceiling-pruned histogram intersection. 
The shortlist is then reranked by BEAM~(Sec.\,\ref{subsubsec:beam}), which restores azimuth information and recovers a coarse yaw estimate through binary bitmask alignment. 
An example is given in Fig.\,\ref{fig:key_construction}\,(b,\,c). 
Finally, the top candidates are passed to geometric verification and pose estimation~(Sec.\,\ref{subsec:verify}), which establishes token-guided correspondences and recovers the full 6-DoF relative pose.

\subsubsection{MINT Shortlisting} \label{subsubsec:mint}

The first stage, termed MINT (\emph{Minimum-ceiling INTersection}), performs efficient database-wide shortlisting using a compact rotation-invariant descriptor derived from the token IDs. It combines pairwise height-ceiling pruning with intersection-based similarity to provide a lightweight first-pass comparison that is robust to heterogeneous VFoV and insensitive to relative yaw.
For each scan, the azimuth component of the token ID is suppressed and the remaining height, radial, and shape indices are accumulated into a histogram $\mathcal{H}\!\in\!\mathbb{R}^{N_h\times(N_rN_s)}$. This produces a height-indexed cylindrical distribution over radial location and local shape, removing horizontal view dependence, which is handled in the next stage.
To account for the differing vertical extents of heterogeneous LiDARs, we apply \emph{height-ceiling pruning}. For a query $\mathcal{T}_{q}$ and database entry $\mathcal{T}_{d}$, the pairwise ceiling is:
\begin{equation} \label{eq:min-height-ceiling}
    \hat{h}_{qd}=\min\big(\max(h^q_b),\max(h^d_b)\big),
\end{equation}
and only rows with $h_b\!\leqslant\!\hat{h}_{qd}$ are retained in both histograms. This avoids penalizing a candidate simply because one sensor observes height layers that are unavailable to the other. 
The surviving rows are summed over height into a compact range-shape vector:
\vspace{-0.45cm}
\begin{equation}
    \mathcal{R}_q = \sum_{h=0}^{\hat{h}_{qd}} \mathcal{H}_q^{(h)},
    \qquad
    \mathcal{R}_d = \sum_{h=0}^{\hat{h}_{qd}} \mathcal{H}_d^{(h)},
\end{equation}
where $\mathcal{R}_{q},\mathcal{R}_{d}\!\in\!\mathbb{R}^{N_rN_s}$. Marginalizing height after ceiling pruning retains the radial and local-shape statistics of the overlapping support while suppressing slice-level sensitivity, yielding a representation suited to fast shortlisting.
The similarity is computed as an L1-normalized histogram intersection:
\vspace{-0.25cm}
\begin{equation}
    S_M(\mathcal{R}_q,\mathcal{R}_d)=
    \frac{\sum_i \min(r_i^q,r_i^d)}
         {\sum_i r_i^q + \epsilon},
\end{equation}
where $r_{i}$ denotes the $i$-th component of the corresponding compact vector $\mathcal{R}$ and $\epsilon$ is a small number for stability. 
The denominator sums over the query, making $S_M$ asymmetric: it measures the fraction of the query distribution contained in the candidate. 
This is well-suited to heterogeneous settings where the query has narrower FoV 
than the database. 
A partial query is not penalized for not spanning the candidate's full distribution. 
The top-$K_1$ candidates move to the next stage.

\subsubsection{BEAM Scoring} \label{subsubsec:beam}
The second stage, termed BEAM (\emph{Binary Elevation-Azimuth Matching}), refines the shortlist by reincorporating azimuth information and estimating the relative yaw between the query and each candidate. 
It converts the tokenized scan into a compact binary elevation-azimuth representation and scores it under circular azimuth shifts.
For a given tokenized scan, we construct a packed binary elevation matrix $\mathcal{A}\!\in\!\{0,\dots,2^{N_h}\!-\!1\}^{N_r\times N_a}$.
Each entry $\mathcal{A}_{r,a}$ encodes the set of occupied height bins in the radial-azimuth cell $(r,a)$ as a binary code, as seen in Fig.~\ref{fig:key_construction}\,(a,\,c):
\begin{equation}
    \mathcal{A}_{r,a} =\!\bigvee_{k \in \mathcal{S}_{r,a}} 2^{h_b^{(k)}},\;\;\;
    \mathcal{S}_{r,a} = \{\,k : r_b^{(k)}\!=\!r,\,a_b^{(k)}\!=\!a\,\}.
\end{equation}
Here, $\vee$ denotes bitwise OR, and $\mathcal{S}_{r,a}$ is the set of tokens assigned to cell $(r,a)$. This collapses all tokens at the same radial-azimuth location into a compact vertical occupancy code, enabling efficient bitwise comparison while preserving the height support required for cross-sensor matching.
As in MINT, comparison is restricted to the overlapping vertical extent: we compute the pairwise ceiling $\hat{h}_{qd}$ from Eq.~\eqref{eq:min-height-ceiling} and mask all BEAM entries so that only bits with $h_b\!\leqslant \!\hat{h}_{qd}$ remain active.
To recover the relative yaw, we score all circular azimuth shifts $n_\theta\!\in\!\{0,\dots,N_a\!-\!1\}$. For each shift, let $\mathcal{A}_d^{(n_\theta)}\!=\!\texttt{roll}(\mathcal{A}_d,n_\theta)$ denote the circularly shifted candidate matrix. The shifted candidate is compared to the query via a bit-level Jaccard similarity:
\vspace{-0.2cm}
\begin{equation}
    S_B(\mathcal{A}_q,\mathcal{A}_d,n_\theta)=
    \frac{\texttt{pop}\left(\mathcal{A}_q \wedge \mathcal{A}_d^{(n_\theta)}\right)}
         {\texttt{pop}\left(\mathcal{A}_q \vee \mathcal{A}_d^{(n_\theta)}\right)},
\end{equation}
where $\wedge$ and $\vee$ denote element-wise bitwise AND and OR, and $\texttt{pop}(\cdot)$ is the total popcount over all matrix entries. Intuitively, this measures the overlap of occupied height bits after compensating for a yaw shift.
To avoid unstable scores from weak overlaps, a shift is considered valid only if the union contains at least $\eta_b$ occupied bits and the aligned pair spans at least $\eta_a$ occupied azimuth columns, evaluated from the column-wise occupancy of the union after shifting. Among valid shifts, the optimal alignment and corresponding yaw estimate are given by:
\vspace{-0.2cm}
\begin{equation}
    n_\theta^* = \arg\max_{n_\theta} S_B(\mathcal{A}_q,\mathcal{A}_d,n_\theta), \;\;
    \hat{\theta}_{qd} = -\,n_\theta^* \frac{2\pi}{N_a}.
\end{equation}
Compared to MINT, BEAM restores directional information and provides a coarse heading estimate while remaining lightweight. Matching is performed on bitmasks via bitwise operations and popcount rather than floating-point histograms. 
The top-$K_2$ candidates together with their associated shifts are passed to the verification stage.

\subsection{Verification and Pose Estimation} \label{subsec:verify}

The final stage takes the top-$K_2$ candidates from BEAM together with their azimuth shifts $n_\theta^*$. It performs token-guided geometric verification with pose estimation, rejecting spurious candidates and recovering a relative pose for those accepted.

\subsubsection{Token-Guided Correspondences} \label{subsubsec:correspondences}

For each candidate, putative 3D correspondences are formed in the ground-aligned frame under the azimuth alignment recovered by BEAM. Given a query token set $\mathcal{T}_q$ and a database token set $\mathcal{T}_d$, the database azimuth bins are circularly shifted by $n_\theta^*$, and correspondences are matched on the spatial indices $(h_b, r_b, a_b)$. Each query token is matched to at most one database token, yielding keypoint index pairs $\mathcal{C}_{qd} = \{(q_k, d_k)\}$.

\subsubsection{6-DoF Pose Estimation} \label{subsubsec:pose}

Pose estimation proceeds in the gravity-aligned frame, where the enforced ground alignment reduces the relative motion to 4-DoF, planar translation $(\hat{t}_x, \hat{t}_y)$, vertical offset $\hat{t}_z$ and yaw $\hat{\theta}$. We apply RANSAC~\cite{RANSAC} to the XY coordinates of $\mathcal{C}_{qd}$ to recover $(\hat{t}_x, \hat{t}_y, \hat{\theta})$ robustly, evaluating inliers by full 3D Euclidean distance.
The surviving inlier set $\mathcal{I}_{qd}\!\subseteq\!\mathcal{C}_{qd}$ is then used to refine the in-plane transform via a closed-form Singular Value Decomposition (SVD) on the centered correspondences, while $\hat{t}_z$ is estimated independently as the median height residual over $\mathcal{I}_{qd}$, which avoids coupling the vertical estimate to the planar optimization and is robust to the sparse and non-uniform height coverage that characterizes heterogeneous submaps. 
To recover the full 6-DoF relative pose in the original sensor frame, the 4-DoF aligned estimate is assembled into the $SE(3)$ transform $\mathbf{T}_\mathrm{aligned}$ and the roll/pitch recovered during ground-plane alignment are reincorporated: 
\begin{equation}
    \mathbf{T}_\mathrm{sensor} =
    \mathbf{T}^{-1}_{\mathrm{ground},d}
    \;\mathbf{T}_\mathrm{aligned}\;
    \mathbf{T}_{\mathrm{ground},q},
\end{equation}
where $\mathbf{T}_{\mathrm{ground},q}$ and $\mathbf{T}_{\mathrm{ground},d}$ are the ground-alignment transforms of the query and database scan, respectively. 

\subsubsection{Verification and Place Acceptance} \label{subsubsec:acceptance}

All query keypoints are transformed by the refined pose, and the ratio of those falling within distance $\delta_\text{inlier}$ of their nearest database counterpart, denoted as the \emph{keypoint inlier ratio} $\rho_\text{inlier}\!\in\![0,1]$, serves as the final verification score. 
A candidate is accepted if $\rho_\text{inlier}\!\geqslant\!\tau$, providing a quality gate before the pose is committed. 

\vspace{-6pt}
\section{Experiments and Results} \label{sec:exp_and_results}
\begin{table*}[!t]
\centering
\caption{Inter-session place recognition results on the HeLiPR dataset. Each method is evaluated across three environments, and seven lidar pairings. Metrics: AUC of the PR curve, Recall@1, and Recall@1\%. \textbf{\green{green}}/\blue{\textbf{blue}}: best and second-best performance.} \label{tab:helipr_results}
\resizebox{\textwidth}{!}{%
\begin{tabular}{cl ccc ccc ccc ccc ccc ccc ccc}
\toprule
& & \multicolumn{6}{c}{\textbf{Identical LiDARs}~~(Database\,$\leftarrow$\,Query)} & \multicolumn{15}{c}{\textbf{Heterogeneous LiDARs}~~(Database\,$\leftarrow$\,Query)}\\
\cmidrule(lr){3-8} \cmidrule(lr){9-23}
& & \multicolumn{3}{c}{{OS2-128\,$\leftarrow$\,OS2-128}}
  & \multicolumn{3}{c}{{Aeries\,II\,$\leftarrow$\,Aeries\,II}}
  & \multicolumn{3}{c}{{OS2-128}\,$\leftarrow$\,VLP-16}
  & \multicolumn{3}{c}{{OS2-128\,$\leftarrow$\,Avia}}
  & \multicolumn{3}{c}{{OS2-128\,$\leftarrow$\,Aeries\,II}}
  & \multicolumn{3}{c}{{Aeries\,II\,$\leftarrow$\,VLP-16}}
  & \multicolumn{3}{c}{{Aeries\,II\,$\leftarrow$\,Avia}} \\
\cmidrule(lr){3-5} \cmidrule(lr){6-8} \cmidrule(lr){9-11} \cmidrule(lr){12-14} \cmidrule(lr){15-17} \cmidrule(lr){18-20} \cmidrule(lr){21-23}
{Seq.} & {Method}
  & AUC & R@1 & R@1\%
  & AUC & R@1 & R@1\%
  & AUC & R@1 & R@1\%
  & AUC & R@1 & R@1\%
  & AUC & R@1 & R@1\%
  & AUC & R@1 & R@1\%
  & AUC & R@1 & R@1\% \\
\midrule

\multirow{7}{*}{\rotatebox{90}{\texttt{Round01-03}}}
& SC++     & 0.86 & 0.67 & 0.91 & 0.70 & 0.58 & 0.90 & 0.02 & 0.01 & 0.08 & 0.01 & 0.01 & 0.02 & 0.01 & 0.01 & 0.02 & 0.01 & 0.01 & 0.12 & 0.06 & 0.06 & 0.30\\
& BTC      & 0.75 & 0.69 & 0.92 & 0.71 & 0.60 & 0.70 & 0.19 & 0.17 & 0.30 & 0.10 & 0.09 & 0.26 & 0.11 & 0.11 & 0.27 & \blue{\textbf{0.08}} & \blue{\textbf{0.08}} & 0.24 & 0.53 & 0.55 & 0.68\\
& SOLiD    & 0.81 & 0.57 & 0.78 & 0.88 & 0.71 & 0.93 & 0.02 & 0.02 & 0.14 & 0.01 & 0.01 & 0.12 & 0.02 & 0.02 & 0.08 & 0.01 & 0.01 & 0.05 & 0.01 & 0.01 & 0.01\\
& HeLiOS   & 0.95 & 0.86 & \green{\textbf{1.00}} & \blue{\textbf{0.98}} & \blue{\textbf{0.91}} & \green{\textbf{0.97}} & \green{\textbf{0.72}} & \green{\textbf{0.66}} & \green{\textbf{1.00}} & 0.10 & 0.10 & 0.58 & 0.13 & 0.13 & \blue{\textbf{0.74}} & 0.02 & 0.02 & \green{\textbf{0.46}} & 0.74 & \blue{\textbf{0.68}} & \green{\textbf{0.96}}\\
& DVMM     & \blue{\textbf{0.99}} & \blue{\textbf{0.88}} & 0.89 & 0.92 & 0.67 & 0.68 & 0.19 & 0.10 & 0.12 & \blue{\textbf{0.80}} & \green{\textbf{0.62}} & \green{\textbf{0.71}} & \blue{\textbf{0.66}} & \blue{\textbf{0.45}} & 0.56 & 0.04 & 0.03 & 0.05 & \blue{\textbf{0.87}} & 0.64 & 0.66\\
\cmidrule(lr){2-23}
& \green{\texttt{\textbf{InLiER}}} & \green{\textbf{0.99}} & \green{\textbf{0.92}} & \blue{\textbf{0.99}} & \green{\textbf{0.99}} & \green{\textbf{0.92}} & \blue{\textbf{0.95}} & \blue{\textbf{0.34}} & \blue{\textbf{0.22}} & \blue{\textbf{0.34}} & \green{\textbf{0.85}} & \blue{\textbf{0.61}} & \blue{\textbf{0.70}} & \green{\textbf{0.86}} & \green{\textbf{0.62}} & \green{\textbf{0.74}} & \green{\textbf{0.32}} & \green{\textbf{0.10}} & \blue{\textbf{0.30}} & \green{\textbf{0.97}} & \green{\textbf{0.81}} & \blue{\textbf{0.87}}\\
\midrule

\multirow{7}{*}{\rotatebox{90}{\texttt{Town01-03}}}
& SC++     & 0.89 & 0.73 & 0.91 & 0.82 & 0.70 & 0.88 & 0.04 & 0.03 & 0.18 & 0.01 & 0.01 & 0.09 & 0.01 & 0.01 & 0.06 & 0.02 & 0.02 & 0.10 & 0.04 & 0.05 & 0.22\\
& BTC      & 0.77 & 0.72 & 0.93 & 0.78 & 0.70 & 0.74 & 0.24 & 0.26 & 0.42 & 0.11 & 0.11 & 0.29 & 0.10 & 0.11 & 0.26 & \blue{\textbf{0.10}} & \blue{\textbf{0.08}} & 0.26 & 0.49 & 0.45 & 0.55\\
& SOLiD    & 0.73 & 0.55 & 0.80 & 0.93 & 0.75 & \blue{\textbf{0.92}} & 0.02 & 0.02 & 0.17 & 0.01 & 0.01 & 0.13 & 0.02 & 0.02 & 0.08 & 0.01 & 0.01 & 0.08 & 0.01 & 0.01 & 0.15\\
& HeLiOS   & 0.97 & \blue{\textbf{0.87}} & \green{\textbf{0.99}} & \green{\textbf{0.98}} & \green{\textbf{0.90}} & \green{\textbf{0.98}} & \green{\textbf{0.76}} & \green{\textbf{0.66}} & \green{\textbf{0.97}} & 0.22 & 0.21 & 0.73 & 0.34 & 0.31 & \green{\textbf{0.82}} & 0.06 & 0.06 & \green{\textbf{0.57}} & \blue{\textbf{0.66}} & 0.58 & \green{\textbf{0.95}}\\
& DVMM     & \blue{\textbf{0.98}} & \blue{\textbf{0.87}} & 0.90 & 0.77 & 0.64 & 0.74 & 0.30 & 0.19 & 0.33 & \blue{\textbf{0.70}} & \blue{\textbf{0.60}} & \green{\textbf{0.80}} & \blue{\textbf{0.67}} & \blue{\textbf{0.53}} & 0.73 & 0.03 & 0.03 & 0.13 & 0.60 & \blue{\textbf{0.59}} & 0.77\\
\cmidrule(lr){2-23}
& \green{\texttt{\textbf{InLiER}}}  & \green{\textbf{0.99}} & \green{\textbf{0.89}} & \blue{\textbf{0.94}} & \blue{\textbf{0.96}} & \blue{\textbf{0.82}} & 0.91 & \blue{\textbf{0.50}} & \blue{\textbf{0.27}} & \blue{\textbf{0.47}} & \green{\textbf{0.87}} & \green{\textbf{0.64}} & \blue{\textbf{0.77}} & \green{\textbf{0.84}} & \green{\textbf{0.63}} & \blue{\textbf{0.79}} & \green{\textbf{0.49}} & \green{\textbf{0.11}} & \blue{\textbf{0.35}} & \green{\textbf{0.67}} & \green{\textbf{0.62}} & \blue{\textbf{0.83}}\\
\midrule

\multirow{7}{*}{\rotatebox{90}{\texttt{Bridge02-03}}}
& SC++     & 0.65 & 0.66 & 0.96 & 0.37 & 0.38 & 0.90 & 0.03 & 0.03 & 0.14 & 0.01 & 0.01 & 0.03 & 0.01 & 0.01 & 0.03 & 0.01 & 0.01 & 0.07 & 0.06 & 0.06 & 0.33\\
& BTC      & 0.63 & 0.62 & 0.73 & 0.59 & 0.53 & 0.61 & 0.23 & 0.22 & 0.41 & 0.10 & 0.11 & 0.20 & 0.10 & 0.10 & 0.19 & \blue{\textbf{0.09}} & \blue{\textbf{0.08}} & 0.19 & 0.33 & 0.32 & 0.38\\
& SOLiD    & 0.31 & 0.27 & 0.58 & 0.56 & 0.54 & 0.88 & 0.02 & 0.02 & 0.11 & 0.01 & 0.01 & 0.05 & 0.01 & 0.01 & 0.04 & 0.01 & 0.01 & 0.03 & 0.01 & 0.01 & 0.06\\
& HeLiOS   & 0.69 & 0.79 & \blue{\textbf{0.98}} & \blue{\textbf{0.72}} & \green{\textbf{0.79}} & \blue{\textbf{0.97}} & \green{\textbf{0.44}} & \green{\textbf{0.50}} & \green{\textbf{0.94}} & 0.08 & \blue{\textbf{0.07}} & 0.68 & 0.13 & 0.14 & \blue{\textbf{0.82}} & 0.04 & 0.04 & \green{\textbf{0.68}} & 0.39 & 0.45 & \green{\textbf{0.95}}\\
& DVMM     & \green{\textbf{0.93}} & \green{\textbf{0.91}} & 0.95 & \green{\textbf{0.76}} & 0.70 & 0.87 & 0.12 & 0.08 & 0.11 & \green{\textbf{0.62}} & \green{\textbf{0.57}} & \green{\textbf{0.80}} & \blue{\textbf{0.48}} & \blue{\textbf{0.36}} & 0.57 & 0.01 & 0.01 & 0.04 & \blue{\textbf{0.52}} & \blue{\textbf{0.56}} & 0.83\\
\cmidrule(lr){2-23}
& \green{\texttt{\textbf{InLiER}}}  & \blue{\textbf{0.76}} & \blue{\textbf{0.89}} & \green{\textbf{0.98}} & 0.66 & \blue{\textbf{0.77}} & \green{\textbf{0.98}} & \blue{\textbf{0.31}} & \blue{\textbf{0.26}} & \blue{\textbf{0.47}} & \blue{\textbf{0.51}} & \blue{\textbf{0.51}} & \blue{\textbf{0.75}} & \green{\textbf{0.54}} & \green{\textbf{0.56}} & \green{\textbf{0.84}} & \green{\textbf{0.28}} & \green{\textbf{0.15}} & \blue{\textbf{0.45}} & \green{\textbf{0.56}} & \green{\textbf{0.59}} & \blue{\textbf{0.92}}\\
\bottomrule

\end{tabular}%
}
\vspace{-16pt}
\end{table*}
We evaluate on the HeLiPR dataset~\cite{minwoo2024helipr}, a heterogeneous inter-session benchmark covering three environments (\texttt{Roundabout}, \texttt{Town}, \texttt{Bridge}) across multiple sessions with four LiDAR types: Ouster OS2-128 ($128$ channels, $360^\circ\!\times\!22.5^\circ$, $200$m), Aeva Aeries\,II (64 channels, $120^\circ\!\times\!19.2^\circ$, $150$m), Velodyne VLP-16 (16 channels, $360^\circ\!\times\!30^\circ$, $100$m), and Livox Avia (6 channels, $70^\circ\!\times\!77^\circ$, $450$m). For full dataset and sensor specifications we refer the reader to~\cite{minwoo2024helipr}. 
Evaluation spans seven sensor pairings: two homogeneous and five cross-sensor, denoted Database\,$\leftarrow$\,Query.
Each submap is formed from $N\!=\!10$ accumulated scans.
InLiER is compared against BTC~\cite{yuan2024btc}, SOLiD~\cite{kim2024solid}, Scan Context++ (SC++)~\cite{kim2022scpp}, HeLiOS~\cite{jung2025helios}, and DVMM~\cite{duan_dvmm_2025}, all executed with their authors' recommended parameters on the same platform (Intel i9-14900K). InLiER is configured with $(N_a, N_r, N_h, N_s)\!=\!(60,20,10,7)$, $v_s\!=\!0.5$m, $[R_{\min},R_{\max}]\!=\![0,100\text{m}]$, $[z_{\min},z_{\max}]\!=\![0,20\text{m}]$, $K_{\max}\!=\!1280$, $K_1\!=\!10\%$ of the database size, and $K_2\!=\!20$.

We report AUC of the precision-recall curve, Recall@1, and Recall@1\% following standard place recognition protocol~\cite{minwoo2024helipr}. 
A candidate is a true positive if $d_\text{TP}\!\leqslant\!d_{\max}\!=\!10$m and $O\!>\!0.2$, where $O$ is the point cloud overlap computed via $0.5$m voxelization following~\cite{jung2025helios}. 
Here $d_\text{TP}$ denotes the ground truth Euclidean distance between a
query and a candidate, and $d_{\max}$ is the maximum $d_\text{TP}$ admitted
as a positive.
The distance threshold is higher than the $3$--$5$m single-scan convention to reflect the larger submap footprint, consistent with~\cite{minwoo2024helipr}. 
The $O\!>\!0.2$ threshold filters geometrically non-informative nearby candidates and is set at the mean overlap of positive pairs under partial-FoV conditions, since higher thresholds yield too few true positives for the Avia and Aeries\,II pairings. 
This joint criterion reflects the operational requirement of SLAM back-ends, where place recognition provides the spatial initialization for metric refinement~\cite{yuan2024btc,Xu2023ringpp}. 
Candidates displaced beyond the convergence basin of a lightweight local aligner yield invalid loop closure constraints regardless of overlap~\cite{stathoulopoulos_frame_2024, yuan2024btc}. Robust global registration can extend this basin at additional cost (Sec.~\ref{subsubsec:distance}), so the criterion keeps retrieval within the regime a lightweight back-end can exploit.
Additional field validation is reported in Section~\ref{subsec:field}.

\vspace{-10pt}
\subsection{LiDAR Place Recognition (LiPR)} \label{subsec:lidar_pr}

\subsubsection{Homogeneous LiPR}

Table~\ref{tab:helipr_results} reports results for the two homogeneous pairings (OS2-128 and Aeries\,II). InLiER achieves the highest AUC and Recall@1 in the majority of configurations, matching or exceeding the closest alternative, across all three environments. Performance is consistent between the \texttt{Roundabout} and \texttt{Town} sequences, while \texttt{Bridge} presents a harder case for all methods due to its geometrically sparse and repetitive structure~\cite{minwoo2024helipr}. In this environment, InLiER maintains competitive Recall@1 and Recall@1\% despite a lower AUC relative to DVMM on the OS2-128 pairing, suggesting that its top-ranked candidates remain reliable even when the overall precision-recall envelope narrows. Notably, InLiER achieves this without any training data, matching the performance of HeLiOS~\cite{jung2025helios} in \texttt{Roundabout} and surpassing it in \texttt{Town} and \texttt{Bridge}.

\vspace{-4pt}
\subsubsection{Heterogeneous LiPR}

The five cross-sensor pairings expose the full heterogeneity challenge. 
SC++ and SOLiD collapse across all heterogeneous configurations. For SC++ this reflects the structured false-signal mode of Section~\ref{sec:introduction}.
SOLiD instead fails because its point-count histograms and per-scan vertical reweighting stay tied to the observing sensor's beam layout and footprint. The descriptor was designed and validated only for a single sensor's restricted FoV, not for cross-sensor asymmetry.
BTC retains modest recall but remains well below InLiER in all pairings. InLiER achieves the best or second-best performance on four of the five cross-sensor pairings, with particularly consistent gains on the OS2-128\,$\leftarrow$\,Avia, OS2-128\,$\leftarrow$\,Aeries\,II, and Aeries\,II\,$\leftarrow$\,Avia configurations across all three environments. 
The VLP-16 pairings are the most demanding, and the two behave differently. 
On OS2-128\,$\leftarrow$\,VLP-16 both sensors are $360^\circ$ spinners sharing broad support, and HeLiOS leads on all metrics.
We attribute this to its training-time exposure to VLP-16 sparsity, which keeps learned features discriminative at 16-channel density where the $z$-span saliency lacks vertically separated returns.
Aeries\,II\,$\leftarrow$\,VLP-16 is different: the rear-mounted, forward-occluded VLP-16 and the front-facing $120^\circ$ Aeries\,II share little instantaneous overlap~\cite{minwoo2024helipr}, and their inter-session co-visible structure is spatially displaced and hard to recover, so AUC and R@1 collapse for every method. 
InLiER still attains the highest AUC in all three environments, which we attribute to submap accumulation thickening this displaced but genuine overlap, whereas HeLiOS's residual edge is confined to R@1\% and reflects its overlap-based retrieval surfacing displaced candidates. 
InLiER outperforms all handcrafted baselines on the VLP-16 pairings and remains the only learning-free method competitive across the full cross-sensor evaluation.

\subsubsection{Retrieval Quality and Loop Closure Utility}
\begin{figure}[!t]
    \centering
    \includegraphics[width=1.0\linewidth]{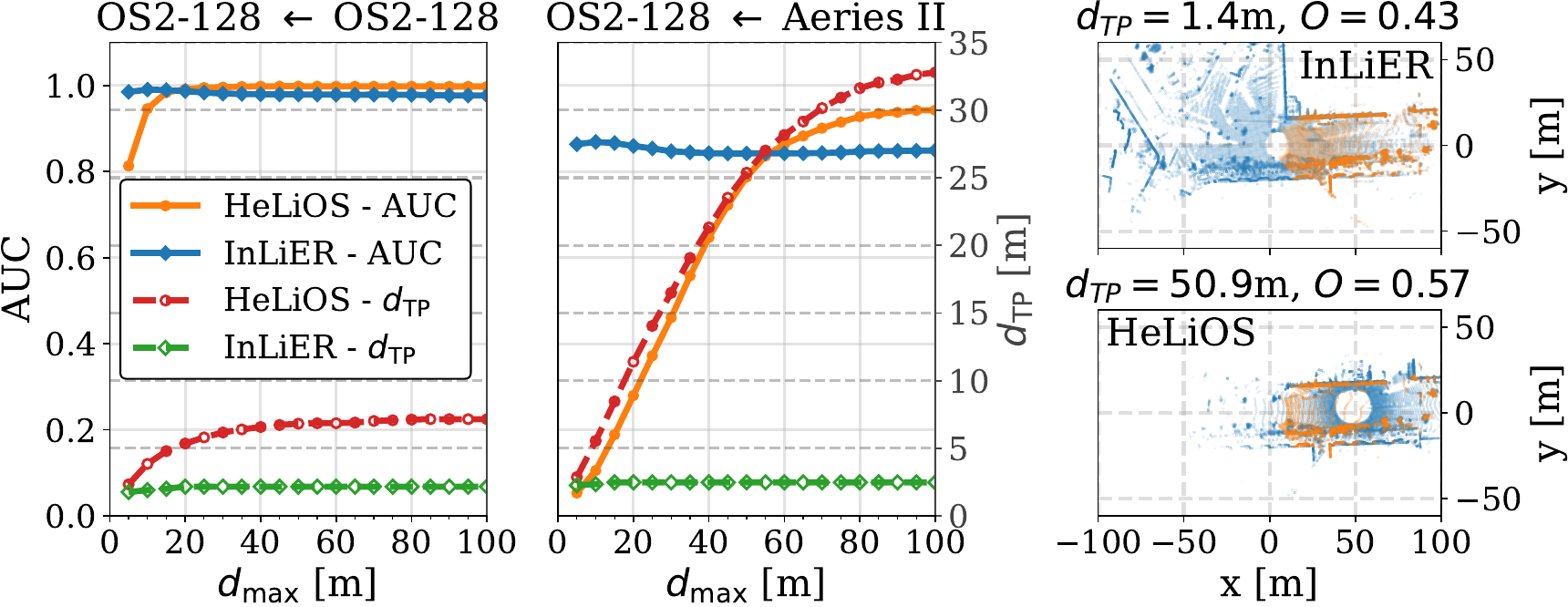}
    \setlength{\abovecaptionskip}{-13pt}
    \caption{Performance as a function of the maximum distance threshold $d_\text{max}$ for InLiER and HeLiOS. The left plot corresponds to a same sensor configuration, while the center plot shows a cross-sensor configuration. Dual axes report AUC (left; \orange{\textbf{orange}}/\blue{\textbf{blue}} solid lines) and the average true positive retrieval distance $d_\text{TP}$ (right; \red{\textbf{red}}/\green{\textbf{green}} dashed lines). The subfigures on the right are an example where for the same query, InLiER retrieves the spatially closest candidate while HeLiOS returns a high-overlap but spatially displaced match.} \label{fig:distance-threshold}
    \vspace{-18pt}
\end{figure}
Place recognition provides a spatial initialization for metric refinement in a SLAM back-end, where ICP-based methods have been shown to fail under large translational offsets~\cite{stathoulopoulos_frame_2024, yuan2024btc, Xu2023ringpp}. 
A high-overlap candidate that is spatially displaced is therefore a poor initialization for a lightweight local aligner
and, absent a robust global registration stage, typically fails to yield a valid loop closure constraint, a failure mode that emerges under
heterogeneous configurations where overlap-based objectives decouple from
spatial proximity.
Fig.~\ref{fig:distance-threshold} illustrates this on \texttt{Roundabout01-03}: as $d_{\max}$ relaxes, HeLiOS's AUC rises as more true positives become available, while InLiER's AUC remains stable across all thresholds, indicating that its top retrievals are already spatially close regardless of how the positive set is defined. $d_\text{TP}$ further confirms this distinction, revealing where the retrieved candidates actually are.
Under the homogeneous configuration, both methods retrieve spatially close candidates across all thresholds, with InLiER at $d_\text{TP}=1.5$m and HeLiOS at $d_\text{TP}=7.5$m. 
Under the cross-sensor configuration, HeLiOS's $d_\text{TP}$ grows substantially with $d_{\max}$. This indicates that its AUC gains come from progressively more displaced retrievals, so its AUC collapses when only spatially close candidates are considered. The pattern suggests that its overlap-based training objective has overfit to high-overlap matches at the expense of spatial proximity.
InLiER's $d_\text{TP}$ remains consistently low in both configurations, reflecting that descriptor similarity in the intermediate token space is grounded in local 3D structure rather than global overlap. Fig.~\ref{fig:distance-threshold} further demonstrates this on a representative retrieval, where HeLiOS returns a candidate at $d_\text{TP}=50.9$m with $O=0.57$, while InLiER retrieves the spatially closest match at $d_\text{TP}=1.4$m with $O=0.43$.

\subsubsection{Metric Pose Estimation}
\begin{table}[!b]
\vspace{-20pt}
\centering
    \caption{Pose estimation on the \texttt{Town01-02} of helipr. te: translational error in meters, re: rotational error in degrees.}\label{tab:pose-estimation}
    \begin{adjustbox}{width=\columnwidth}
    \begin{tabular}{c c c c c c c}
    \toprule
    \multirow{2}{*}{Pair} & \multicolumn{2}{c}{\green{\texttt{\textbf{InLiER}}}} & \multicolumn{2}{c}{\green{\texttt{\textbf{InLiER}}} w. $\mathcal{K}$--GICP}  & \multicolumn{2}{c}{\green{\texttt{\textbf{InLiER}}} w. $\mathcal{P}$--GICP} \\
    \cmidrule(lr){2-3} \cmidrule(lr){4-5} \cmidrule(lr){6-7}
     & {\textbf{TE} [m]} & {\textbf{RE} [$^\circ$]} & \textbf{TE} [m] & {\textbf{RE} [$^\circ$]} & \textbf{TE} [m] & {\textbf{RE} [$^\circ$]}\\
    \midrule
    \multirow{2}{*}{OS2-128\,$\leftarrow$\,OS2-128} & 1.024 & 1.189 & 0.290 & 0.437 & 0.229 & 0.168\\
    & $\pm$0.676 & $\pm$0.797 & $\pm$ 0.276 & $\pm$0.303 & $\pm$0.237 & $\pm$0.177\\
    \midrule
    \multirow{2}{*}{OS2-128\,$\leftarrow$\,Aeries\,II} & 1.140 & 2.143 & 0.655 & 1.256 & 0.277 & 0.250\\
    & $\pm$0.529 & $\pm$1.342 & $\pm$0.390 & $\pm$1.022 & $\pm$0.157 & $\pm$0.146\\
    \midrule
    \multirow{2}{*}{Aeries\,II\,$\leftarrow$\,Aeries\,II} & 1.179 & 1.750 & 0.617 & 0.951 & 0.226 & 0.128\\
    & $\pm$0.600 & $\pm$0.937 & $\pm$0.489 & $\pm$0.740 & $\pm$0.230 & $\pm$0.074\\
    \bottomrule
    \end{tabular}
    \end{adjustbox}
    \vspace{1pt}
    \raggedright
{\footnotesize \\ 
$\mathcal{K}$\,/\,$\mathcal{P}$--GICP: Pose refinement with the keypoints or raw points, respectively. \\
}
\end{table}
Table~\ref{tab:pose-estimation} reports pose estimation results on \texttt{Town01-02} across homogeneous and cross-sensor configurations. 
The standalone token-guided pose achieves translational errors below $1.2$m and rotational errors below $2.2^\circ$, with only a minor degradation under the cross-sensor pairing, confirming that the intermediate token representation preserves sufficient geometric structure for metric pose recovery even under sensor heterogeneity. 
This result is particularly relevant in bandwidth-constrained or multi-agent scenarios, where the compact $K_{\max}\!\times\!1$ token representation can be transmitted directly and a valid pose estimate recovered without access to the raw point cloud. 
Keypoint-level refinement with $\mathcal{K}$-GICP yields a substantial improvement at modest additional cost, since matching operates on the already-available $K_{\max}\!\times\!4$ keypoint set rather than the full point cloud. 
Full refinement with $\mathcal{P}$-GICP achieves the lowest errors across all configurations, and remains available when maximum accuracy is required. 
The three operating modes offer a natural accuracy-transmission tradeoff, from compact token-only pose up to full point cloud refinement, without altering the retrieval pipeline.
\begin{figure}[!b]
    \centering
    \vspace{-8pt}
    \includegraphics[width=1.0\linewidth]{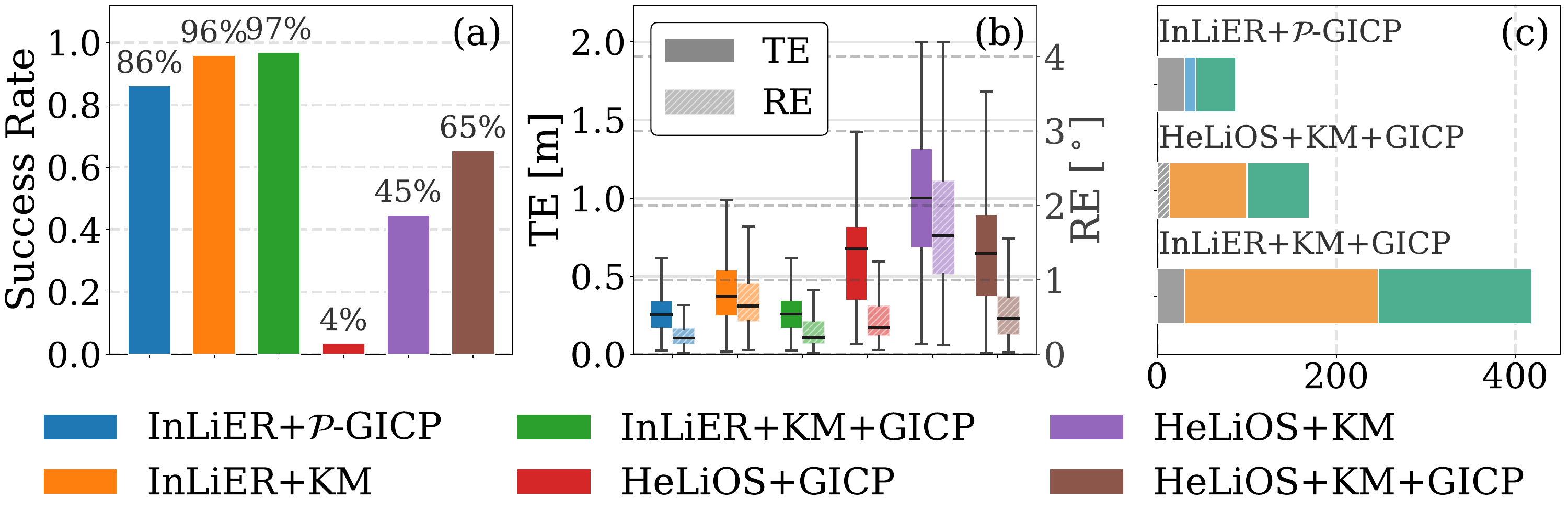}
    \setlength{\abovecaptionskip}{-14pt}    
    \caption{Registration study at $d_{\max}\!=\!60$m on \texttt{Roundabout01-03} (OS2-128\,$\leftarrow$\,Aeries\,II). Top-1 retrievals are aligned with GICP, KISS-Matcher (KM), or combined. $\mathcal{P}$-GICP is seeded by InLiER's pose estimation. Success requires (TE,\,RE)$\,\leqslant\!(2\text{m},\,5^\circ)$. (a)~Success rate. (b)~Translation and rotation error over successful pairs. (c)~Mean per-query runtime (HeLiOS on GPU).}\label{fig:backend}
\end{figure}

\vspace{-12pt}
\subsubsection{Registration Back-End Comparison} \label{subsubsec:distance}
Our joint distance-overlap true positive criterion presumes that a spatially displaced candidate lies outside the convergence basin of a local aligner and thus can yield an invalid loop closure constraint.
This premise is specific to local registration since robust global aligners such as KISS-Matcher (KM)~\cite{lim2025kissmatcher} are built to absorb large misalignment and could, in principle, make spatial proximity unnecessary. 
We therefore test whether the distance term of our criterion is still required once a robust global aligner is placed in the back-end.
At $d_{\max}\!=\!60$m, the threshold where InLiER and HeLiOS reach equal retrieval AUC in Fig.~\ref{fig:distance-threshold}, we register each method's top-1 retrievals with GICP~\cite{small_gicp}, KM, and KM+GICP, declaring a success when $(\text{TE},\,\text{RE})\!\leqslant\!(2\text{m},\,5^\circ)$~\cite{lim2025kissmatcher}.
Fig.~\ref{fig:backend}\,(a) shows InLiER's built-in prior with $\mathcal{P}$-GICP registering $86\%$ of retrievals with no global stage, against $4\%$ for HeLiOS+GICP. 
KM raises both ($96\%$ vs. $45\%$), confirming that the back-end is a dominant factor. 
The ceiling for HeLiOS+KM, however, shows that a robust aligner still leaves its more displaced retrievals frequently unregistered.
Over the pairs each configuration registers, InLiER+$\mathcal{P}$-GICP attains the lowest median error, below the three-stage HeLiOS+KM+GICP, as depicted in Fig.~\ref{fig:backend}\,(b), while remaining fastest at $87$ms versus $170$ms and $418$ms for the KM pipelines as shown in Fig.~\ref{fig:backend}\,(c). 
This holds even though InLiER runs entirely on CPU while HeLiOS's descriptor extraction requires a GPU.
Our joint distance-overlap criterion is therefore a design choice that keeps retrieval within the convergence basin a lightweight GICP back-end can exploit, avoiding a dedicated global registration or outlier rejection stage.

\vspace{-10pt}
\subsection{Ablation Studies and Runtime} \label{subsec:ablations}
\begin{figure}[!t]
    \centering
    \includegraphics[width=1.0\linewidth]{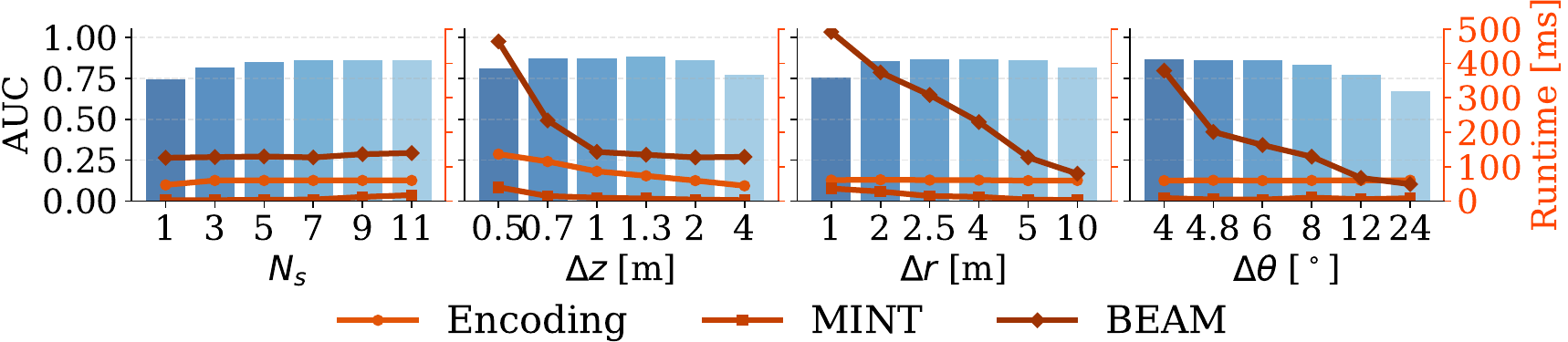}
    \setlength{\abovecaptionskip}{-15pt}
    \caption{Parameter sensitivity and per-module runtime for InLiER. AUC (left axis) and runtime (right axis) are reported as a function of $N_s,\,\Delta z,\,\Delta r,\,\Delta\theta$.} \label{fig:ablations}
    \vspace{-16pt} 
\end{figure}

Fig.~\ref{fig:ablations} examines the sensitivity of InLiER to its four key parameters, evaluated on OS2-128. Increasing $N_s$ from $1$ to $7$ yields consistent AUC improvements, confirming that shape classification provides a meaningful discriminative signal beyond spatial binning alone. Beyond $N_s\!=\!7$ performance plateaus with negligible runtime impact, as finer angular subdivisions of the shape classes yield diminishing returns in urban environments where structural primitives are dominated by horizontal and vertical configurations. For the resolution parameters, derived as $\Delta z\!=\!(z_{\max}-z_{\min})/N_h$, $\Delta r\!=\!R_{\max}/N_r$, and $\Delta\theta\!=\!2\pi/N_a$, AUC degrades at coarse discretizations where distinct structures collapse into the same bin, while finer resolutions plateau without further gain. 
The defaults $(N_h, N_r, N_a)\!=\!(10,20,60)$, corresponding to $(\Delta z, \Delta r, \Delta\theta)\!=\!(2\text{m},5\text{m},6^\circ)$, 
balance discriminability and compactness. 
Per-query Python runtimes at default parameters: encoding $70$ms, MINT $4.5$ms, and BEAM $160$ms. Encoding is the worst case due to the high density of the OS2-128; the sparser Aeries\,II yields $40$ms. MINT stays lightweight by operating on marginalized histograms, while BEAM dominates due to the exhaustive circular shift scoring over $N_a$ azimuth bins. 

\subsection{Field Experiment Validation} \label{subsec:field}

Field validation is conducted on a campus dataset collected with a Clearpath Warthog UGV across two $\sim\!2.7$km sessions, each yielding $\sim\!1000$\,submaps per session. 
The database is built from an Ouster OS0-32 (32 spinning beams, $360^\circ\!\times\!90^\circ$, $75$m range) and queries are collected with a Seyond Robin-W (semi-solid-state, 192 scan lines, $120^\circ\!\times\!70^\circ$, $150$m range), challenging cross-modal pairing in FoV, scanning modality, and point density. 
The two sessions overlap only partially, further increasing retrieval difficulty.
\begin{figure}[!t]
    \centering
    \includegraphics[width=1.0\linewidth]{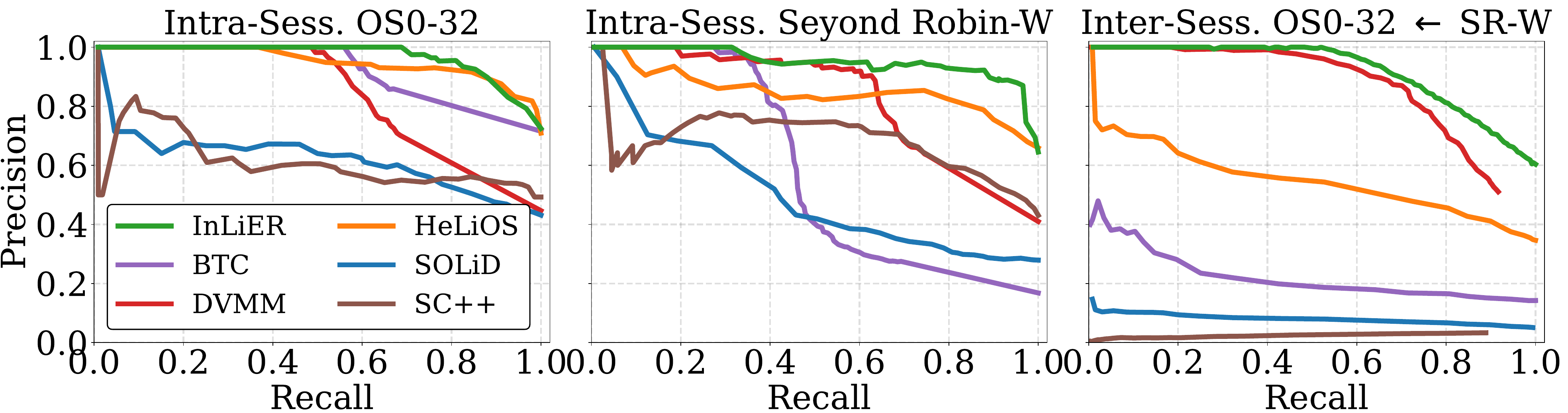}
    \setlength{\abovecaptionskip}{-12pt}
    \caption{Precision-recall curves for the compared methods in the campus field experiment. Results are shown for intra-session place recognition with the Ouster OS0-32 and Seyond Robin-W (SR-W) sensors, and for inter-session retrieval from Robin-W queries to the OS0-32 database. InLiER maintains high precision across all three settings, showing the most consistent performance under both homogeneous and heterogeneous LiDAR conditions.}
    \vspace{-8pt}
    \label{fig:campus_auc}
\end{figure}
\begin{figure}[!t]
    \centering
    \includegraphics[width=1.0\linewidth]{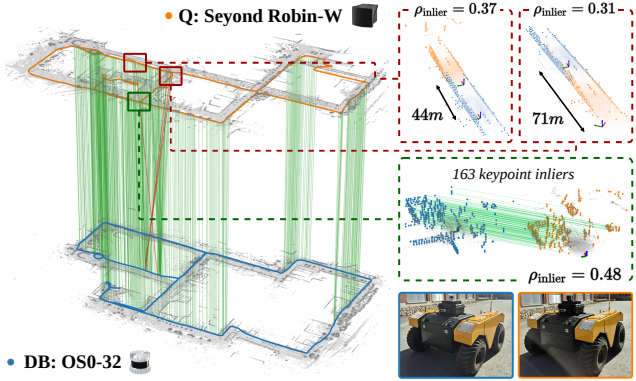}
    \setlength{\abovecaptionskip}{-10pt}
    \caption{Loop closure edges from the campus field experiment using an Ouster OS0-32 as the database (DB) sensor and a Seyond Robin-W as the query (Q) sensor, mounted on a Clearpath Warthog UGV. True positives are shown in \green{\textbf{green}} and false positives in \red{\textbf{red}}, with candidates accepted for $\rho_\text{inlier} \!\geqslant\!0.3$. The false positives arise from self-similar plain wall segments, where scans separated by $44$m and $71$m are incorrectly aligned yielding high inlier ratios.} \label{fig:campus_loop_closures}
    \vspace{-15pt} 
\end{figure}

Fig.~\ref{fig:campus_auc} reports the precision-recall curves (authors' default parameters; HeLiOS evaluated zero-shot from its released weights) across three configurations. 
InLiER achieves an AUC of $0.96$, $0.94$, and $0.91$ for intra-session OS0-32, intra-session Robin-W, and inter-session OS0-32\,$\leftarrow$\,SR-W respectively, outperforming all baselines in each setting, with the largest margin on the heterogeneous pairing.
DVMM is also competitive, likely because both sensors are mounted at similar heights, satisfying its shared equatorial FoV assumption~\cite{duan_dvmm_2025}. 
Fig.~\ref{fig:campus_loop_closures} shows the accepted loop closures at $\rho_\text{inlier}\!\geqslant\!0.3$, yielding 400 true positives and only 4 false positives concentrated in a structurally self-similar plain wall region. 
The 200 false negatives (against 400 true positives and 300 true negatives) arise primarily in areas of high pedestrian density: slow-moving individuals are registered across accumulated submaps, introducing spurious keypoints that reduce descriptor consistency. 
Fast-moving vehicles leave sparser, less disruptive artifacts.
This identifies dynamic obstacles during submap accumulation as a current limitation. 
The OS0-32 database submaps average 360\,keypoints ($1.4$KB token-only; $9.8$KB with keypoints), the narrower-FoV Robin-W queries 230\,keypoints ($0.9$KB; $6.2$KB), keeping the token set well within practical transmission budgets for multi-agent deployments.


\section{Conclusions and Future Work} \label{sec:conclusions}

We presented InLiER, a learning-free global localization pipeline for heterogeneous LiPR that maps scans into a shared vocabulary of height-sliced keypoints and mixed-radix tokens derived solely from local 3D geometry, decoupling place description from sensor FoV, resolution, and scanning pattern. 
On the HeLiPR benchmark across seven sensor pairings and in a real-world cross-sensor field experiment, InLiER achieves state-of-the-art performance among handcrafted methods and outperforms the learning-based baseline on most cross-sensor configurations. 
It retrieves spatially close candidates suitable for SLAM loop closure, requires no training, runs on CPU, and uses a sub-2KB representation.
Its main limitation is sensitivity to dynamic obstacles, particularly slow pedestrians that create spurious keypoints during submap accumulation.
Future work will address dynamic point removal, integrate InLiER with SLAM back-ends, and explore informed keyframe~\cite{stathoulopoulos2026msa} and submap selection~\cite{thorne2025submod} to reduce map redundancy.

\bibliographystyle{./IEEEtranBST/IEEEtran}
\bibliography{./IEEEtranBST/IEEEabrv,root}

@ARTICLE{yuan2024btc,
  author={Yuan, Chongjian and Lin, Jiarong and Liu, Zheng and Wei, Hairuo and Hong, Xiaoping and Zhang, Fu},
  journal={IEEE Transactions on Robotics}, 
  title={{BTC: A Binary and Triangle Combined Descriptor for 3-D Place Recognition}}, 
  year={2024},
  volume={40},
  number={},
  pages={1580-1599},
  doi={10.1109/TRO.2024.3353076}
}

@ARTICLE{Xu2023ringpp,
  author={Xu, Xuecheng and Lu, Sha and Wu, Jun and Lu, Haojian and Zhu, Qiuguo and Liao, Yiyi and Xiong, Rong and Wang, Yue},
  journal={IEEE Transactions on Robotics}, 
  title={{RING++: Roto-Translation Invariant Gram for Global Localization on a Sparse Scan Map}}, 
  year={2023},
  volume={39},
  number={6},
  pages={4616-4635},
  doi={10.1109/TRO.2023.3303035}
}

@INPROCEEDINGS{jung2025helios,
  author={Jung, Minwoo and Jung, Sangwoo and Gil, Hyeonjae and Kim, Ayoung},
  booktitle={2025 IEEE International Conference on Robotics and Automation (ICRA)}, 
  title={{HeLiOS: Heterogeneous Lidar Place Recognition via Overlap-Based Learning and Local Spherical Transformer}}, 
  year={2025},
  volume={},
  number={},
  pages={2204-2211},
  doi={10.1109/ICRA55743.2025.11127677}
}

@ARTICLE{kim2024solid,
  author={Kim, Hogyun and Choi, Jiwon and Sim, Taehu and Kim, Giseop and Cho, Younggun},
  journal={IEEE Robotics and Automation Letters}, 
  title={{Narrowing Your FOV With SOLiD: Spatially Organized and Lightweight Global Descriptor for FOV-Constrained LiDAR Place Recognition}}, 
  year={2024},
  volume={9},
  number={11},
  pages={9645-9652},
  doi={10.1109/LRA.2024.3440089}
}

@ARTICLE{kim2022scpp,
  author={Kim, Giseop and Choi, Sunwook and Kim, Ayoung},
  journal={IEEE Transactions on Robotics}, 
  title={{Scan Context++: Structural Place Recognition Robust to Rotation and Lateral Variations in Urban Environments}}, 
  year={2022},
  volume={38},
  number={3},
  pages={1856-1874},
  doi={10.1109/TRO.2021.3116424}
}

@article{minwoo2024helipr,
  author = {Minwoo Jung and Wooseong Yang and Dongjae Lee and Hyeonjae Gil and Giseop Kim and Ayoung Kim},
  title ={{HeLiPR: Heterogeneous LiDAR dataset for inter-LiDAR place recognition under spatiotemporal variations}},
  journal = {The International Journal of Robotics Research},
  volume = {43},
  number = {12},
  pages = {1867-1883},
  year = {2024},
  doi = {10.1177/02783649241242136}
}

@article{duan_dvmm_2025,
    title = {{DVMM}: {A} {Dual}-{View} {Combination} {Descriptor} for {Multi}-{Modal} {LiDARs} {Online} {Place} {Recognition}},
    volume = {10},
    copyright = {https://ieeexplore.ieee.org/Xplorehelp/downloads/license-information/IEEE.html},
    issn = {2377-3766, 2377-3774},
    shorttitle = {{DVMM}},
    doi = {10.1109/LRA.2025.3600141},
    language = {en},
    number = {10},
    urldate = {2026-01-21},
    journal = {IEEE Robotics and Automation Letters},
    author = {Duan, Xuzhe and Hu, Qingwu and Ai, Mingyao and Zhao, Pengcheng and Li, Jiayuan},
    month = oct,
    year = {2025},
    pages = {10434--10441},
}

@article{stathoulopoulos_frame_2024,
    title = {{FRAME}: {A} {Modular} {Framework} for {Autonomous} {Map} {Merging}: {Advancements} in the {Field}},
    volume = {1},
    copyright = {https://ieeexplore.ieee.org/Xplorehelp/downloads/license-information/IEEE.html},
    issn = {2997-1101},
    shorttitle = {{FRAME}},
    doi = {10.1109/TFR.2024.3419439},
    language = {en},
    number = {1},
    urldate = {2024-11-21},
    journal = {IEEE Transactions on Field Robotics},
    author = {Stathoulopoulos, Nikolaos and Lindqvist, Björn and Koval, Anton and Agha-Mohammadi, Ali-Akbar and Nikolakopoulos, George},
    year = {2024},
    pages = {1--26},
}

@article{RANSAC,
author = {Fischler, Martin A. and Bolles, Robert C.},
title = {{Random Sample Consensus: A Paradigm for Model Fitting with Applications to Image Analysis and Automated Cartography}},
year = {1981},
issue_date = {June 1981},
publisher = {Association for Computing Machinery},
address = {New York, NY, USA},
volume = {24},
number = {6},
issn = {0001-0782},
doi = {10.1145/358669.358692},
journal = {Commun. ACM},
month = jun,
pages = {381–395},
numpages = {15}
}

@article{weinmann2015semantic,
title = {{Semantic Point Cloud Interpretation based on Optimal Neighborhoods, Relevant Features and Efficient Classifiers}},
journal = {ISPRS Journal of Photogrammetry and Remote Sensing},
volume = {105},
pages = {286-304},
year = {2015},
issn = {0924-2716},
doi = {https://doi.org/10.1016/j.isprsjprs.2015.01.016},
author = {Martin Weinmann and Boris Jutzi and Stefan Hinz and Clément Mallet},
}

@article{pca,
title = {{Principal Components Analysis (PCA)}},
journal = {Computers \& Geosciences},
volume = {19},
number = {3},
pages = {303-342},
year = {1993},
issn = {0098-3004},
doi = {https://doi.org/10.1016/0098-3004(93)90090-R},
author = {Andrzej Maćkiewicz and Waldemar Ratajczak},
}

@article{small_gicp,
author = {Kenji Koide},
title = {{small\_gicp: Efficient and Parallel Algorithms for Point Cloud Registration}},
journal = {Journal of Open Source Software},
month = aug,
number = {100},
pages = {6948},
volume = {9},
year = {2024},
doi = {10.21105/joss.06948}
}

@article{Luo2024pcpcsurvey,
author={Luo, Kan
and Yu, Hongshan
and Chen, Xieyuanli
and Yang, Zhengeng
and Wang, Jingwen
and Cheng, Panfei
and Mian, Ajmal},
title={{3D Point Cloud-based Place Recognition: A Survey}},
journal={Artificial Intelligence Review},
year={2024},
month={Mar},
day={07},
volume={57},
number={4},
pages={83},
issn={1573-7462},
doi={10.1007/s10462-024-10713-6},
}

@ARTICLE{loopclosures,
  author={Tsintotas, Konstantinos A. and Bampis, Loukas and Gasteratos, Antonios},
  journal={IEEE Transactions on Intelligent Transportation Systems}, 
  title={{The Revisiting Problem in Simultaneous Localization and Mapping: A Survey on Visual Loop Closure Detection}}, 
  year={2022},
  volume={23},
  number={11},
  pages={19929-19953},
  doi={10.1109/TITS.2022.3175656}
}

@ARTICLE{reloc,
  author={Shi, Chenghao and Chen, Xieyuanli and Xiao, Junhao and Dai, Bin and Lu, Huimin},
  journal={IEEE Transactions on Robotics}, 
  title={{Fast and Accurate Deep Loop Closing and Relocalization for Reliable LiDAR SLAM}}, 
  year={2024},
  volume={40},
  number={},
  pages={2620-2640},
  doi={10.1109/TRO.2024.3386363}
}

@ARTICLE{stathoulopoulos2026msa,
  author={Stathoulopoulos, Nikolaos and Kanellakis, Christoforos and Nikolakopoulos, George},
  journal={IEEE Robotics and Automation Letters}, 
  title={{A Minimal Subset Approach for Informed Keyframe Sampling in Large-Scale SLAM}}, 
  year={2026},
  volume={11},
  number={1},
  pages={738-745},
  doi={10.1109/LRA.2025.3636035}}

@INPROCEEDINGS{thorne2025submod,
  author={Thorne, David and Chan, Nathan and Ma, Yanlong and Robison, Christa S. and Osteen, Philip R. and Lopez, Brett T.},
  booktitle={2025 IEEE International Conference on Robotics and Automation (ICRA)}, 
  title={{Submodular Optimization for Keyframe Selection \& Usage in SLAM}}, 
  year={2025},
  volume={},
  number={},
  pages={5033-5039},
  doi={10.1109/ICRA55743.2025.11127770}}

@misc{jung_treeloc_2026,
    title = {{TreeLoc}: 6-{DoF} {LiDAR} {Global} {Localization} in {Forests} via {Inter}-{Tree} {Geometric} {Matching}},
    shorttitle = {{TreeLoc}},
    doi = {10.48550/arXiv.2602.01501},
    language = {en},
    urldate = {2026-03-10},
    publisher = {arXiv},
    author = {Jung, Minwoo and Chebrolu, Nived and Lima, Lucas Carvalho de and Oh, Haedam and Fallon, Maurice and Kim, Ayoung},
    month = feb,
    year = {2026},
    note = {arXiv:2602.01501 [cs]},
}

@INPROCEEDINGS{lim2025kissmatcher,
  author={Lim, Hyungtae and Kim, Daebeom and Shin, Gunhee and Shi, Jingnan and Vizzo, Ignacio and Myung, Hyun and Park, Jaesik and Carlone, Luca},
  booktitle={2025 IEEE International Conference on Robotics and Automation (ICRA)}, 
  title={{KISS-Matcher: Fast and Robust Point Cloud Registration Revisited}}, 
  year={2025},
  volume={},
  number={},
  pages={11104-11111},
  doi={10.1109/ICRA55743.2025.11127458}
}

@INPROCEEDINGS{gupta2024efficiently,
  author={Gupta, Saurabh and Guadagnino, Tiziano and Mersch, Benedikt and Vizzo, Ignacio and Stachniss, Cyrill},
  booktitle={2024 IEEE International Conference on Robotics and Automation (ICRA)}, 
  title={{Effectively Detecting Loop Closures using Point Cloud Density Maps}}, 
  year={2024},
  volume={},
  number={},
  pages={10260-10266},
  doi={10.1109/ICRA57147.2024.10610962}
}


\end{document}